# Structured Message Passing

**Vibhav Gogate**
Department of Computer Science
The University of Texas at Dallas
Richardson, TX 75080, USA.
vgogate@hlt.utdallas.edu

**Pedro Domingos**
Computer Science & Engineering
University of Washington
Seattle, WA 98195, USA.
pedrod@cs.washington.edu

## Abstract

In this paper, we present structured message passing (SMP), a unifying framework for approximate inference algorithms that take advantage of structured representations such as algebraic decision diagrams and sparse hash tables. These representations can yield significant time and space savings over the conventional tabular representation when the message has several identical values (context-specific independence) or zeros (determinism) or both in its range. Therefore, in order to fully exploit the power of structured representations, we propose to artificially introduce context-specific independence and determinism in the messages. This yields a new class of powerful approximate inference algorithms which includes popular algorithms such as cluster-graph Belief propagation (BP), expectation propagation and particle BP as special cases. We show that our new algorithms introduce several interesting bias-variance trade-offs. We evaluate these trade-offs empirically and demonstrate that our new algorithms are more accurate and scalable than state-of-the-art techniques.

## 1 INTRODUCTION

Access to fast, scalable and accurate approximate inference algorithms is the key to the successful application of graphical models to real world problems. As a result, several approximate inference algorithms have been proposed to date, in a large body of literature spanning several decades. Existing algorithms can be classified into two broad types: message passing based and sampling or simulation based. Message passing algorithms operate by passing messages over the edges of a cluster graph derived from the graphical model while sampling algorithms operate by randomly generating variable configurations. In this paper, we focus on message passing algorithms and propose a new framework, *structured message passing* (SMP) which provides a principled approach for taking advantage of structured approaches for representing and manipulating messages.

We propose SMP because popular, approximate message passing algorithms such as belief propagation (BP) [22], its various generalizations [19, 34], and expectation propagation (EP) [20, 21] rely on tabular representations. Tabular representations, although easy to use and manipulate, can be exponentially worse in terms of size and processing time than structured approaches such as algebraic decision diagrams (ADDs) [1] and sparse hash tables. As a result, in presence of time and space resource constraints, which is often the case in practice, we are unable to apply several more efficient and potentially more accurate classes of algorithms to real-world problems.

Over the last decade, there has been much research on developing exact inference algorithms that exploit the power of structured representations. Notable examples are Cachet [25], ACE [4], and ADD-based variable elimination [3]. The first two use weighted propositional features for representing messages while ADD-based variable elimination uses ADDs [1]. By taking advantage of structural features such as context-specific independence (CSI) [2] and determinism, the aforementioned algorithms can solve much larger problems than the junction tree algorithm [17]. For approximate inference, however, structured representations have not been investigated as much (cf. [5, 10, 18, 27]) and their power has not been fully realized.

The basic idea in SMP is quite simple. Unlike BP and EP in which we associate each cluster and each edge in a cluster graph with a single tabular function and a product of tabular functions respectively [33], in SMP we associate a structured representation of a function with each cluster and each edge, yielding a *structured cluster graph*. We assume that the structured representation not only defines a suitable (computer) representation but also various inference operators such as sum and product. Thus, given a cluster graph and a message passing schedule, each representation defines a structured message passing algorithm.

We show that in spite of its simplicity, SMP enables us to define more powerful BP and EP algorithms as well as several new classes of (principled) message passing algorithms. In particular, when the inference operators are lossless, i.e., they faithfully represent the message, we get the cluster graph BP algorithm. When the inference operators are lossy and minimize the KL divergence between the original function and the lossy representation, we get the EP algorithm. Defining new lossy operators yields new classes of algorithms. However, since the structured representations can be exponentially more efficient than the tabular representation, the resulting SMP algorithms are likely to be much more efficient in terms of time and space complexity. Thus, given a bound on time and space complexity, SMP will allow much larger clusters than tabular BP and EP. Since the accuracy typically increases with the cluster size, it is likely that SMP algorithms will be more accurate than tabular BP and EP.

We consider a possible instance of the class of SMP algorithms in which we artificially introduce determinism and CSI in the messages. Such messages can then be efficiently represented using structured approaches, yielding a significant reduction in complexity. Moreover, if each new message includes assignments that have relatively high information content, the resulting algorithm will also have high accuracy. We propose to introduce determinism via Monte Carlo simulation (e.g., via Gibbs sampling or importance sampling), retaining only the sampled (and therefore potentially high-probability) partial assignments within each cluster. Following [10, 30], we propose to introduce CSI by quantizing messages, namely reducing the number of distinct values in the range of the message by replacing a number of values that are close to each other by a single value.

We show that our new SMP algorithm introduces several bias-variance trade-offs. Specifically, we show that given a set of samples and a fixed error bound for quantization, increasing the cluster size increases the variance but reduces the bias. On the other hand, for a fixed error bound and cluster size, increasing the sample size decreases the variance and therefore improves accuracy.

Within our algorithm and the SMP framework, we consider two structured representations: sparse hash tables and algebraic decision diagrams,[1] define lossy and lossless operators for them and empirically evaluate their efficacy on various benchmarks. For comparison, we use iterative join graph propagation (IJGP) [19], co-winner of 2010 UAI competition [8], and evaluate our algorithms on the task of computing all single variable marginal distributions. Our experiments show that our new algorithm is superior to IJGP.

The rest of the paper is organized as follows. In section 2, we describe notation and preliminaries. In section 3, we introduce structured cluster graphs and describe structured representations and operators in section 4. In section 5, we present our new algorithm that is a possible instance of SMP and analyze bias-variance tradeoffs for it. We empirically evaluate our new algorithm in section 6. We discuss related work in section 7 and conclude in section 8.

## 2 PRELIMINARIES AND NOTATION

A (discrete) graphical model or a Markov network (cf. [6, 15, 24]), denoted by $\mathcal{G}$, is a triple $(\mathbf{X}, \mathbf{D}, \Phi)$, where $\mathbf{X} = \{X_1, \ldots, X_n\}$ is a set of variables, $\mathbf{D} = \{D(X_1), \ldots, D(X_n)\}$ is a set of domains of variables, where $D(X_i)$ is the domain of $X_i$ and $\Phi = \{\phi_1, \ldots, \phi_m\}$ is a set of functions (also called factors or potentials). Each function $\phi_i$ is defined over a subset of variables, called its scope, denoted by $S(\phi_i)$. Let $D(\mathbf{X})$ denote the Cartesian product of the domains of all variables in $\mathbf{X}$. Let $\mathbf{x} = (x_1, \ldots, x_n) \in D(\mathbf{X})$ where $x_i \in D(X_i)$ denote an assignment of values to all variables in $\mathbf{X}$. A Markov network represents the following probability distribution.

$$P_{\mathcal{G}}(\mathbf{x}) = \frac{\prod_{i=1}^{m} \phi_i(\mathbf{x}_{S(\phi_i)})}{\sum_{\mathbf{x} \in D(\mathbf{X})} \prod_{i=1}^{m} \phi_i(\mathbf{x}_{S(\phi_i)})} \quad (1)$$

where $\mathbf{x}_{S(\phi_i)}$ is the projection of $\mathbf{x}$ on $S(\phi_i)$. We will often abuse notation and write $\phi_i(\mathbf{x}_{S(\phi_i)})$ as $\phi_i(\mathbf{x})$. The denominator of Eq. (1) is a normalization constant, called the *partition function*. Common queries over graphical models are computing the partition function and the marginal distribution $P_{\mathcal{G}}(X_i)$ for all variables $X_i \in \mathbf{X}$.

Cluster graph belief propagation (BP) is an approximate message passing algorithm for computing variable marginals. It operates on a data structure called the cluster graph defined below:

**Definition 1.** Given a graphical model $\mathcal{G} = (\mathbf{X}, \mathbf{D}, \Phi)$, a **cluster graph** is a graph $G(\mathbf{V}, \mathbf{E})$ in which each vertex $V \in \mathbf{V}$ and edge $E \in \mathbf{E}$ is associated with a subset of variables, denoted by $L(V)$ and $L(E)$ respectively such that: (i) for every function $\phi \in \Phi$, there exists a vertex $L(V)$ such that $S(\phi) \subseteq L(V)$; and (ii) for every variable $X \in \mathbf{X}$, the set of vertices and edges in $G$ that mention $X$ form a connected sub-tree of $G$ (*running intersection property*).

In cluster graph BP, we first put each function $\phi \in \Phi$ in a cluster $V$ such that $S(\phi) \subseteq L(V)$. Then each node $V_i$ sends the following message to a node $V_j$ on the edge $E_{i,j}$, iteratively until convergence

$$m_{i \to j}(\mathbf{y}) = \sum_{\mathbf{z}} \prod_{\phi \in \Phi(V_i)} \phi(\mathbf{y}, \mathbf{z}) \prod_{V_k \in N(i,j) \setminus \{V_j\}} m_{k \to i}(\mathbf{y}, \mathbf{z})$$

[1] Note that SMP is a general approach for easily designing message-passing algorithms and as such can be used with any structured representation, not just ADDs and sparse hash tables. For example, it is relatively straight-forward to extend our basic framework to Affine ADDs [26].

where $\mathbf{Y} = L(E_{i,j})$, $\mathbf{y} \in D(\mathbf{Y})$, $\mathbf{Z} = L(V_i) \setminus L(E_{i,j})$, $\mathbf{z} \in D(\mathbf{Z})$, $\Phi(V_i)$ is the set of functions from the graphical model assigned to $V_i$, $m_{i\to j}$ is the message sent from $V_i$ to $V_j$, and $N(i,j)$ is the set of neighbors of $V_i$ in $G$. Once the messages have converged, we can recover the marginal distribution for any variable $X_i \in \mathbf{X}$ by finding a cluster $V \in \mathbf{V}$ that mentions $X_i$, multiplying all functions and incoming messages to the cluster and then summing out all variables other than $X_i$ from the resulting function. The cluster graph BP algorithm may not converge. In such cases, we can put a bound on the number of iterations and stop the algorithm once it exceeds this bound.

The message passing approach described above is called sum-product message passing. An alternative approach, which has smaller time complexity but higher space complexity is belief-update message passing (see [15] for more details). When the cluster graph is a tree, cluster graph BP is exact and coincides with the junction tree algorithm. The time and space complexity of cluster graph BP is $O(|\mathbf{V}| \exp(\max_{V\in\mathbf{V}} |L(V)|))$ assuming that each message is represented using a table.

The expectation propagation algorithm (EP) [20] operates on a cluster graph by passing approximate messages. In EP, we associate a product of functions with each edge [33], denoted by $\widetilde{m}_{i\to j} = \prod_k \widetilde{m}_{i\to j,k}$. The main idea here is to approximate a large message which is computationally infeasible using a tractable message $\widetilde{m}_{i\to j}$ such that the KL divergence[2] between the two is minimized.

Unlike cluster graph BP, in EP, sum-product and belief-update message passing will yield different estimates. Often, however, we prefer belief-update message passing in EP because it yields more accurate answers in practice.

# 3 STRUCTURED CLUSTER GRAPHS

In a cluster graph, each cluster and each edge is associated with a tabular function or a product of tabular functions. The main, fairly simple idea in structured cluster graphs is to associate each edge and each cluster with a parametric (i.e., structured) representation of a function. The parametric representation of a function is a pair $(\mathcal{R}, \mathbf{w})$ where $\mathcal{R}$ denotes the structure and $\mathbf{w}$ is a set of real-valued parameters. We assume throughout that $\mathcal{R}$ determines the complexity of representing the function. We also assume that the structure is fixed or we have bound on its complexity.

**Definition 2.** Given a graphical model $(\mathbf{X}, \mathbf{D}, \Phi)$, a **structured cluster graph** (SCG) is a graph $G(\mathbf{V}, \mathbf{E})$ in which each vertex $V \in \mathbf{V}$ and each edge $E \in \mathbf{E}$ is associated with a parametric representation of a function, denoted by $\mathcal{R}_V$ and $\mathcal{R}_E$ respectively, such that: (i) for every function $\phi \in \Phi$, there exists a vertex $V$ such that $S(\phi) \subseteq S(\mathcal{R}_V)$ and (ii) for every variable $X \in \mathbf{X}$, the set of vertices and edges in $G$ that mention $X$ form a connected sub-tree of $G$.

[2] KL divergence between two distributions $P$ and $Q$ is given by $\sum_{\mathbf{x}} P(\mathbf{x}) \log(P(\mathbf{x})/Q(\mathbf{x}))$. To compute KL divergence between two functions, we normalize the functions and then compute KL divergence between them.

To perform message passing over a SCG, we need to define the sum, product and division operators over the parametric representation. We assume that the *representation system* used defines these operators for us. In addition, we assume that the system provides a *projection* operator, which takes a function $\phi$ and a parametric representation $(\mathcal{R}, \mathbf{w})$ as input and sets the parameters $\mathbf{w}$. In other words, the projection operator yields an instantiation of $(\mathcal{R}, \mathbf{w})$, which we will denote by $\mathcal{R}[\phi]$. We say that $\mathcal{R}[\phi]$ is *lossless* if we can recover $\phi(\mathbf{y})$ for all $\mathbf{y} \in D(S(\phi))$, i.e., $\mathcal{R}[\phi](\mathbf{y}) = \phi(\mathbf{y})$. Otherwise, it is *lossy*.

The *product* and *division* operators take two instantiations $\mathcal{R}_A[\phi_i]$ and $\mathcal{R}_B[\phi_j]$, and a target representation $\mathcal{R}_C$ as input, and output $\mathcal{R}_C[\phi_k]$ where $\phi_k = \phi_i.\phi_j$ and $\phi_k = \phi_i/\phi_j$ respectively. The *sum* operator takes as input an instantiation $\mathcal{R}_A[\phi_i]$, a representation $\mathcal{R}_B$, and a set of variables $\mathbf{Y} \subseteq S(\phi_i)$ as input and outputs $\mathcal{R}_B[\phi_j]$ where $\phi_j = \sum_{\mathbf{Y}} \phi_i$. We say that the sum, product and division operators are lossless if their output is lossless. Otherwise, they are lossy. The lossy sum, product and division operators can be defined in terms of their lossless analogues using the projection operator; the lossy instantiation $\mathcal{R}_{LY}[\phi]$, is simply a projection of the lossless instantiation $\mathcal{R}_{LS}[\phi]$, on $\mathcal{R}_{LY}$.

The message passing algorithm over a SCG, which we will refer to as *structured message passing* (SMP) operates as follows. First, we initialize the SCG by initializing the parametric representation at each edge and each cluster to the uniform distribution (or to some other distribution based on prior knowledge). Then, for each function $\phi \in \Phi$, we select a cluster $V$ such that $S(\phi) \subseteq L(V)$ and multiply $\mathcal{R}_V[\phi]$ with the current structured representation, say $\mathcal{R}_V[\phi_V]$ at $V$, storing the result in $\mathcal{R}_V[\phi_V]$. Then, we pass messages, as usual, between the clusters, using the sum, division and product operators, until convergence. In sum-product propagation only the sum and product operators are used while in the belief update propagation all the three operators are used. The complexity of structured message passing is dependent on the representation system used.

It is straight-forward to show that:

**Proposition 1.** *SMP is equivalent to the cluster graph BP algorithm if all operators are lossless. Similarly, SMP is equivalent to the EP algorithm under the restriction that the sum, product and division operators are lossy and all messages on all edges $E_{i,j} = (V_i, V_j)$ are such that they minimize the K-L divergence between the actual message $m_{i\to j}$ and the represented message $\mathcal{R}_{E_{i,j}}[m_{i\to j}]$.*

In spite of this equivalence, note that SMP with the afore-

| A | B | $\phi(a,b)$ |
|---|---|---|
| 0 | 0 | 3 |
| 0 | 1 | 3 |
| 1 | 0 | 5 |
| 1 | 1 | 0 |

(a)

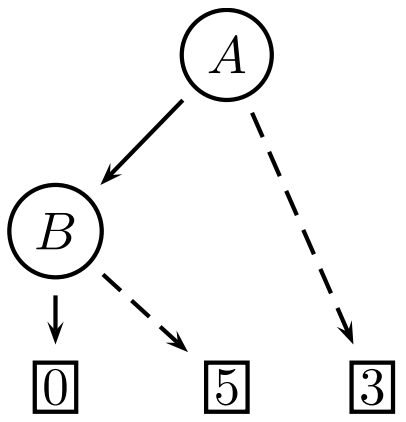

(b)

Figure 1: (a) Tabular representation of a Boolean function, (b) Its ADD representation. In the ADD representation the solid and dashed arcs correspond to the true and false assignments of the parent variable respectively.

mentioned restrictions is more powerful than both tabular BP and tabular EP because by using structured representations that are more efficient than tabular representations, in practice, we can run SMP on graphs having much larger cluster size than both BP and EP. As the accuracy typically increases with the cluster size, we expect SMP to be more accurate than tabular EP and BP algorithms having comparable computational complexity.

## 4 STRUCTURED REPRESENTATIONS

In this section, we consider two structured representations, sparse tables and algebraic decision diagrams, and describe lossless sum, product, division and assignment operators for them. Then, we show how we can exploit the power of these representations by defining lossy operators.

**Sparse Tables** Sparse or zero-suppressed tables are often useful when a substantial number of zeros are present in the graphical model [13, 16]. Instead of storing a real number for all possible configurations of variables, the function is represented as a list of tuples having non-zero values. For example, the sparse representation of the tabular representation given in Fig. 1(a) is a table that contains only the first three entries. The non-zero tuples are typically stored in a hash table for fast access. It is easy to define lossless sum, product, division and assignment operators over this representation. For instance, the product operator corresponds to database hash join and the sum operator corresponds to database project (cf. [28]). The complexity of these operations is linear in the size of the input and output tables, assuming constant time table lookup. The lossless operators define a cluster graph BP algorithm, which is likely to be more efficient than the tabular approach when determinism is present. However, when no determinism is present, the resulting BP algorithm will be slightly inefficient than the tabular approach because of the constant time overhead introduced by the sparse operators.

**Algebraic Decision Diagrams** Algebraic decision diagrams (ADDs) [1] are an efficient representation of real-valued Boolean functions having many identical values in their range. ADDs are directed acyclic graphs (DAG) and have two types of nodes: leaf nodes which are labeled by real-values and non-leaf nodes which are labeled by variables. Each decision node has two outgoing arcs corresponding to the true and false assignments of the corresponding variable. ADDs enforce a strict variable ordering from the root to the leaf node and impose the following three constraints on the DAG: (i) no two arcs emanating from a decision node can point to the same node, (ii) if two decision nodes have the same variable label, then they cannot have (both) the same true child node and the same false child node and (iii) no two leaf nodes are labeled by the same real value. ADDs that do not satisfy these constraints are referred to as unreduced ADDs while those that do are called reduced ADDs. An unreduced ADD can be reduced by merging isomorphic subgraphs and eliminating any nodes whose two children are isomorphic (see [1] for more details). A reduced, ordered ADD is a canonical representation. Namely, two functions will have the same ADD (under the same variable ordering) iff they are the same. For example, Fig. 1(b) shows the ADD representation of the function given in Fig. 1(a).

It is easy to define lossless sum, product and division operators using standard ADD operations (and in practice, implement them using open-source ADD packages such as CUDD [29]). The complexity of these operations is linear in the size of the largest ADD. Note that any non-Boolean function can be converted to a Boolean function by introducing a Boolean variable for each variable-value pair and adding Boolean constraints which ensure that each variable is assigned exactly one value (cf. [32]). Therefore, our approach is also applicable to multi-valued variables.

**ADDs and Sparse Tables as Features** ADDs and sparse tables can be interpreted as representations of weighted features (or propositional formulas) defined over their variables. Each entry in the sparse table represents a simple conjunctive feature while each leaf node of an ADD represents a complex feature that is a disjunction of several conjunctive features. For example, the first two entries in the sparse table in Fig. 1(a) represent the conjunctive weighted features $[(\neg A \wedge \neg B), 3]$ and $[(\neg A \wedge B), 3]$ respectively while the rightmost leaf node in the ADD in Fig. 1(b) represents the complex weighted feature $[((\neg A \wedge \neg B) \vee (\neg A \wedge B)), 3]$, which is logically equivalent to $[(\neg A), 3]$.

### 4.1 Lossy Operators

In order to fully exploit the power of structured representations, we need lossy operators. Note that without lossy operators, we cannot guarantee that the size of the computed message will be bounded by the size of the structure associated with each edge. Since lossy sum, product and division operators are simply lossy projections of their lossless counterparts, we only have to define the lossy projection operator.

**Definition 3.** Given a ADD (or sparse table) $\mathcal{R}$, let $[f_i, v_i]$ be the weighted feature associated with a leaf node (entry) $i$. Then, the **lossy projection** of a probability distribution $\phi$ on $i$ is $v_i = \frac{1}{|\texttt{Sol}(f_i)|} \sum_{\mathbf{y} \in \texttt{Sol}(f_i)} \phi(\mathbf{y})$ where $\texttt{Sol}(f_i)$ is the set of assignments that are consistent with $f_i$. The **lossy** projection of $\phi$ on $\mathcal{R}$, denoted by $\mathcal{R}[\phi]$ is the lossy projection of $\phi$ on all leaf nodes (entries) of $\mathcal{R}$.

We can show that:

**Theorem 1.** *Given a ADD (or sparse table) $\mathcal{R}$, the lossy projection operator (see Definition 3) minimizes the KL divergence between $\phi$ and $\mathcal{R}[\phi]$. In other words, there exists no other ADD (or sparse table) that has the same structure as $\mathcal{R}$ but has smaller KL divergence.*

At first glance, the lossy projection operator may seem impractical because it involves computing the number of assignments that are consistent with a feature, i.e., it includes solving the #$\mathcal{P}$-complete model counting problem. However, for sparse hash tables and ADDs, this is not an issue because model counting is constant time and linear time in the size of the representation respectively.

## 5 A STRUCTURED MESSAGE PASSING ALGORITHM

Clearly, in order to exploit the compactness of ADDs and sparse tables, a majority of the messages should contain determinism and/or CSI. To this end, we propose to artificially introduce CSI and determinism in the messages. Intuitively, if the introduced CSI and determinism captures most of the probability mass in the message, the resulting algorithm will be as accurate as cluster graph BP. However, its time and space complexity will be much smaller.

We cannot add zeros or determinism arbitrarily, however. Notice that in order to guarantee that the KL divergence between the exact message and all possible projected messages does not equal infinity, all configurations of variables that are consistent in the exact message should also be consistent in the projected message. Otherwise, the minimum KL distance will equal infinity. For example,

**Example 1.** Consider a cluster that is associated with an ADD representing two features $(A \vee B \vee C, 5)$ and $(\neg A \wedge \neg B \wedge \neg C, 0)$ and an edge with its neighbor that is associated with an ADD representing two features $(\neg A \wedge \neg B, v)$ and $(A \vee B, 0)$. No matter what value of $v$ is selected, the KL divergence between the exact message obtained by summing out $C$ from the cluster and any message projected on the structured representation will be infinity.

There are a number of ways in which we can add determinism so that all clusters and edges satisfy the consistency condition described above. We propose the following approach because of its simplicity: generate a set of samples and project them on each cluster and each edge. The projected samples define a constraint that all partial assignments that are not present in the generated samples have zero weight. For example,

**Example 2.** Consider a graphical model with three binary variables $\{X_1, X_2, X_3\}$. Let $\mathbf{S} = \{(0,1,1),(1,0,0),(1,0,1)\}$ be a set of samples over the three variables. Consider a cluster defined over two variables $\{X_1, X_2\}$. Then, the projection of $\mathbf{S}$ on the cluster is the relation: $\{(0,1),(1,0)\}$. The other two assignments $\{(1,1),(0,0)\}$ have zero weight. The ADD associated with this cluster will represent the following set of features: $\{(\neg X_1 \wedge X_2, v_1)$, $(X_2 \wedge \neg X_2, v_2)$, $((\neg X_1 \wedge \neg X_2) \vee (X_1 \wedge X_2), 0)\}$.

We can show that our approach that introduces determinism via sampling is correct and yields a structured EP algorithm. The only assumption we have to make is that each tuple having non-zero value in each function in the graphical model is included in the samples. This will ensure that the KL divergence between any function in the graphical model and its lossy projection is finite. Formally,

**Theorem 2.** *Given a graphical model $\mathcal{G} = (\mathbf{X}, \mathbf{D}, \Phi)$, a structured cluster graph $G(\mathbf{V}, \mathbf{E})$, let $\mathbf{S}$ be a set of samples over $\mathbf{X}$ such that: (1) For each cluster $V \in \mathbf{V}$ and each edge $E \in \mathbf{E}$, $\mathcal{R}_V$ and $\mathcal{R}_E$ are such that for any function $\phi$ and for all configurations $\mathbf{x} \notin \mathbf{S}$, $\mathcal{R}_V[\phi](\mathbf{x}) = 0$ and $\mathcal{R}_E[\phi](\mathbf{x}) = 0$ and for all configurations $\mathbf{x} \in \mathbf{S}$, $\mathcal{R}_V[\phi](\mathbf{x}) > 0$ and $\mathcal{R}_E[\phi](\mathbf{x}) > 0$ and (2) For each function $\phi \in \Phi$, all assignments $\mathbf{x} \in S(\phi)$ such that $\phi(\mathbf{x}) > 0$ are included in $\mathbf{S}$. Then, for each edge $E$, there exists a lossy message such that the KL divergence between the lossless message and the lossy one is finite.*

To artificially introduce CSI in the message, we propose to use quantization [10, 30]. In this approach, given a small real number $\epsilon$, we put all values in the range of the function into multiple bins such that the absolute difference between any two values in each bin is bounded by $\epsilon$. The goal is to minimize the number of bins. Then, we replace all values in each bin by their average value in each function. Quantization reduces the number of distinct values in the range of the function and as a result reduces the size of the ADD representing the message.

The discussion above yields Algorithm 1, which is a possible instance of SMP. The algorithm takes as input a graphical model $\mathcal{G} = (\mathbf{X}, \mathbf{D}, \Phi)$, a cluster graph $G(\mathbf{V}, \mathbf{E})$, a representation system $\mathcal{R}$, an integer $k$ denoting the number of samples and a real number $\epsilon$, which denotes the error bound used for quantization. The algorithm first generates samples from the graphical model. The samples can be generated using either importance sampling or Gibbs sampling. (For higher accuracy, the samples should be such that they

**Algorithm 1:** Structured Message Passing

**Input**: A graphical model $\mathcal{G} = (\mathbf{X}, \mathbf{D}, \Phi)$, a cluster graph $G = (\mathbf{V}, \mathbf{E})$, a representation system $\mathcal{R}$, integer $k > 0$ and a real number $0 \leq \epsilon \leq 1$
**Output**: A structured cluster graph with (converged) messages and potentials
**begin**
  **S**=Generate $k$ Samples from $\mathcal{G}$;
  **for** *each* $V \in \mathbf{V}$ *and* $E \in \mathbf{E}$ **do**
    `// Project` $\mathbf{S}$ `on` $V$ `and` $E$
    Initialize $\mathcal{R}_V[\phi_V]$ and $\mathcal{R}_E[\phi_E]$ to zero;
    **for** *all configurations* $\mathbf{x} \in \mathbf{S}$ **do**
      set $\mathcal{R}_V[\phi_V](\mathbf{x}) = 1$ and $\mathcal{R}_E[\phi_E](\mathbf{x}) = 1$;
  Let $G' = (\mathbf{V}', \mathbf{E}')$ be the structured cluster graph obtained in the above step;
  **for** *each function* $\phi$ *in* $\Phi$ **do**
    Find a cluster $V \in \mathbf{V}'$ such that $S(\phi) \subseteq S(\phi_V)$ and multiply $\phi$ with $\mathcal{R}_V[\phi_V]$;
  Run sum-product or belief-update message passing on $G'$ until convergence. Quantize each message using $\epsilon$;

capture the modes of the distribution.) After the samples are generated, we project the samples on each cluster and each edge and use the representation system to yield a structured cluster graph $G'(\mathbf{V}', \mathbf{E}')$. Then, we initialize the parameters of each cluster $V \in \mathbf{V}'$ and each edge $E \in \mathbf{E}'$ using the functions in the graphical model. Finally, the algorithm runs either sum-product or belief-update message passing on the structured cluster graph. Each message is quantized using $\epsilon$.

Algorithm 1 is equivalent to the cluster graph BP algorithm when $k = \infty$ and $\epsilon = 0$. It is equivalent to the quantization-based EP algorithm proposed in [10] when $k = \infty$ and $\epsilon > 0$. For other values of $k$ and $\epsilon$, Algorithm 1 yields a Monte Carlo approximation of cluster graph BP and EP.

The algorithm just presented belongs to a class of algorithms that combine sampling-based inference with message-passing based inference. Many other advanced algorithms proposed in literature such as Particle BP [14, 31], AND/OR sampling [9], and sample propagation [23] belong to this class. The novelty in our proposed algorithm is that we combine sampling-based inference with message-passing inference over structured cluster-graphs. This yields several interesting bias versus variance tradeoffs, which can be leveraged to improve the accuracy of estimation. We discuss these tradeoffs next.

### 5.1 Analysis: Bias-Variance Tradeoffs

In this section, we analyze the bias-variance tradeoffs in Algorithm 1. Let $f(\mathbf{x})$ be the quantity that we want to estimate (e.g., the partition function or the posterior marginals). Given a set of samples $\mathbf{S}$, a cluster graph $G$ and constant $\epsilon$, let $h(\mathbf{x}; G, \epsilon, \mathbf{S})$ denote the estimate of $f(\mathbf{x})$ computed using Algorithm 1 with $G, \mathbf{S}, \epsilon$ as inputs. Then, the expected mean squared error between $f(\mathbf{x})$ and $h(\mathbf{x}; G, \epsilon, \mathbf{S})$ is

$$\mathbb{E}_{\mathbf{S}}[\{h(\mathbf{x}; G, \epsilon, \mathbf{S}) - f(\mathbf{x})\}^2] = \{\mathbb{E}_{\mathbf{S}}[h(\mathbf{x}; G, \epsilon, \mathbf{S})] - f(\mathbf{x})\}^2 + \{\mathbb{E}_{\mathbf{S}}[\{h(\mathbf{x}; G, \epsilon, \mathbf{S}) - \mathbb{E}_{\mathbf{S}}[h(\mathbf{x}; G, \epsilon, \mathbf{S})]\}^2]\} \quad (2)$$

The first term in Eq. (2) equals bias squared and the second term equals the variance.

We can show that:

**Theorem 3.** *Increasing the cluster size (or decreasing $\epsilon$) of the cluster graph used by Algorithm 1 decreases the asymptotic bias $\lim_{|\mathbf{S}| \to \infty} |\mathbb{E}_{\mathbf{S}}[h(\mathbf{x}; G, \epsilon = 0, \mathbf{S})] - f(\mathbf{x})|$ but increases the variance.*

*Proof.* (Sketch) Notice that in the limit of infinite samples and assuming that $\epsilon = 0$, Algorithm 1 is equivalent to the cluster graph BP algorithm (we also assume that the sampling algorithm generates every assignment having non-zero probability in $\mathcal{G}$ with non-zero probability). Since the set of cluster graphs whose cluster size is bounded by $i$ (along with the associated cluster potentials and messages) is included in the set of cluster graphs whose cluster size is bounded by $i+1$, the bias will never increase as we increase the cluster size. Moreover, cardinality arguments (the number of different clusters of size $i+1$ is far greater than the number of different clusters of size $i$) dictate that there exists a particular setting of cluster potentials and messages in a cluster graph whose cluster size is bounded by $i+1$ that cannot be represented by any cluster graph whose cluster size is bounded by $i$ and therefore the asymptotic bias decreases as we increase the cluster size. (The asymptotic bias of a junction tree is zero).

To prove that the variance increases as we increase the cluster size, consider two clusters $V$ and $V'$ where $V'$ is constructed from $V$ by adding a variable to it. Clearly, projecting the given set $\mathbf{S}$ of samples on $V$ will cover a greater percentage of the tuples in the potential associated with $V$ as compared to the potential associated with $V'$. As a result, the effective sample size at $V$ is larger than the effective sample size at $V'$. Since the variance decreases as the sample size increases, the variance at $V$ will be smaller than the variance at $V'$. Therefore, the variance increases as we increase the cluster size. Increasing $\epsilon$ has the effect of introducing new context specific (conditional) independences, which is the same as decreasing the cluster size. Therefore, the same arguments apply to decreasing $\epsilon$. □

From the central limit theorem, it is immediate that:

**Theorem 4.** *Increasing the number of samples while fixing $G$ and $\epsilon$ decreases the variance. Moreover, the sample bias converges to the asymptotic bias at the rate of $O(|\mathbf{S}|^{-1/2})$.*

Theorems 3 and 4 summarize the bias-variance tradeoffs associated with Algorithm 1. For a fixed sample size, cluster graphs having large clusters will typically have low bias and high variance while cluster graphs having small clusters will typically have large bias and low variance.

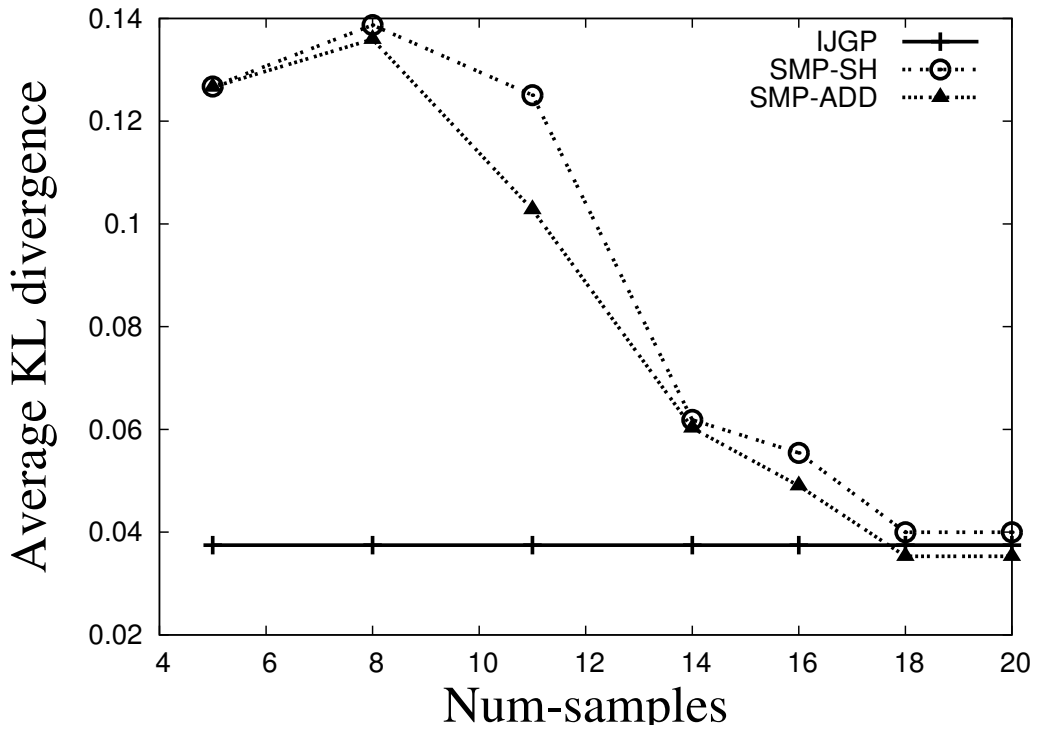

(a)

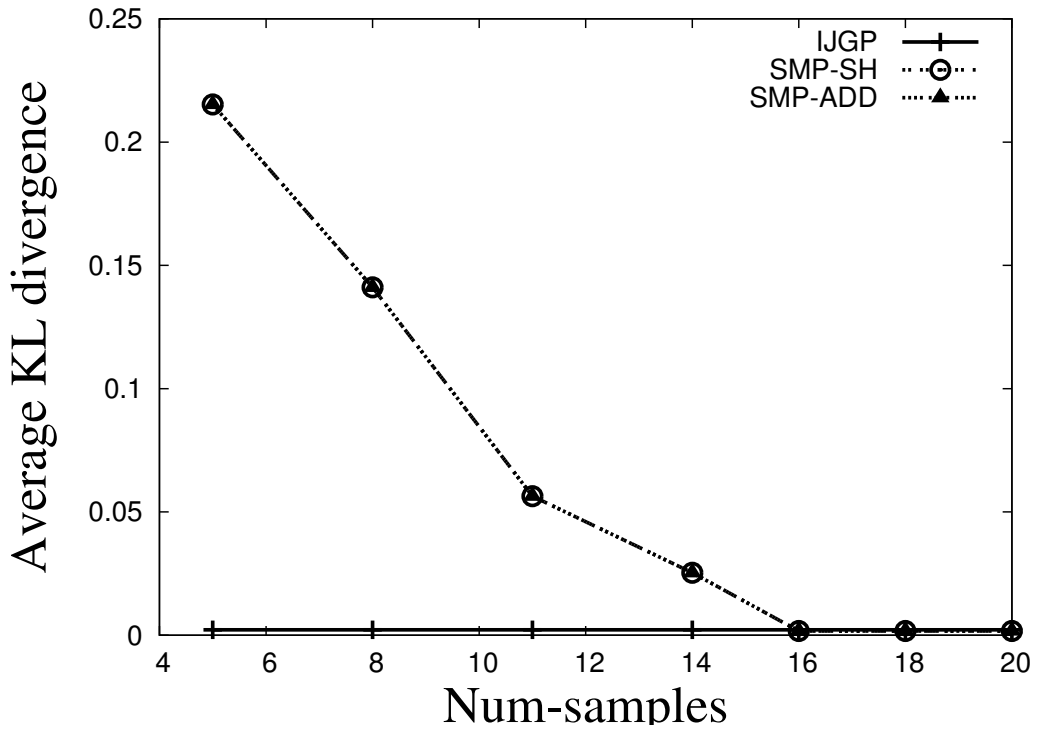

(b)

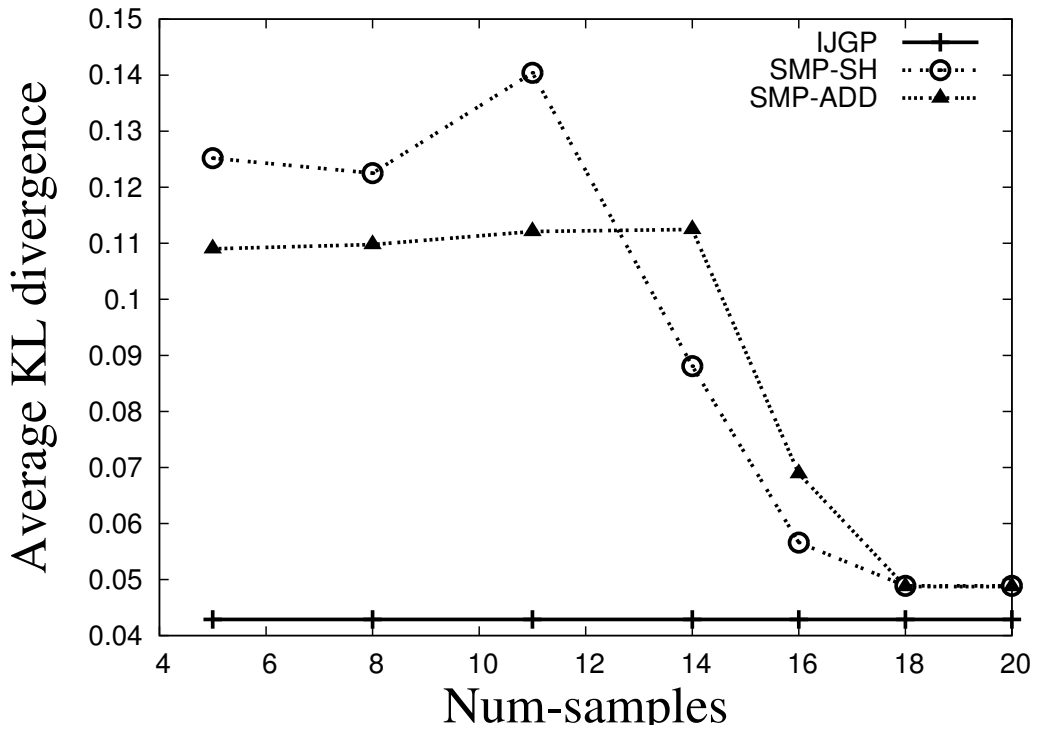

(c)

Figure 2: Impact of varying $k$ (number of samples), keeping the i-bound and $\epsilon$ constant for (a) $20 \times 20$ Ising model with 10 evidence nodes, $i = 9$ and $\epsilon = 2^{-40}$, (b) Block coding instance with 255 variables and 511 functions, $i = 12$ and $\epsilon = 2^{-100}$ and (c) Logistics instance with 1413 nodes and 29487 functions, $i = 15$ and $\epsilon = 2^{-40}$.

# 6 EXPERIMENTS

In this section, we compare SMP with Iterative Join Graph propagation (IJGP) [19], a state-of-the-art tabular cluster graph BP algorithm. IJGP won two out of the three

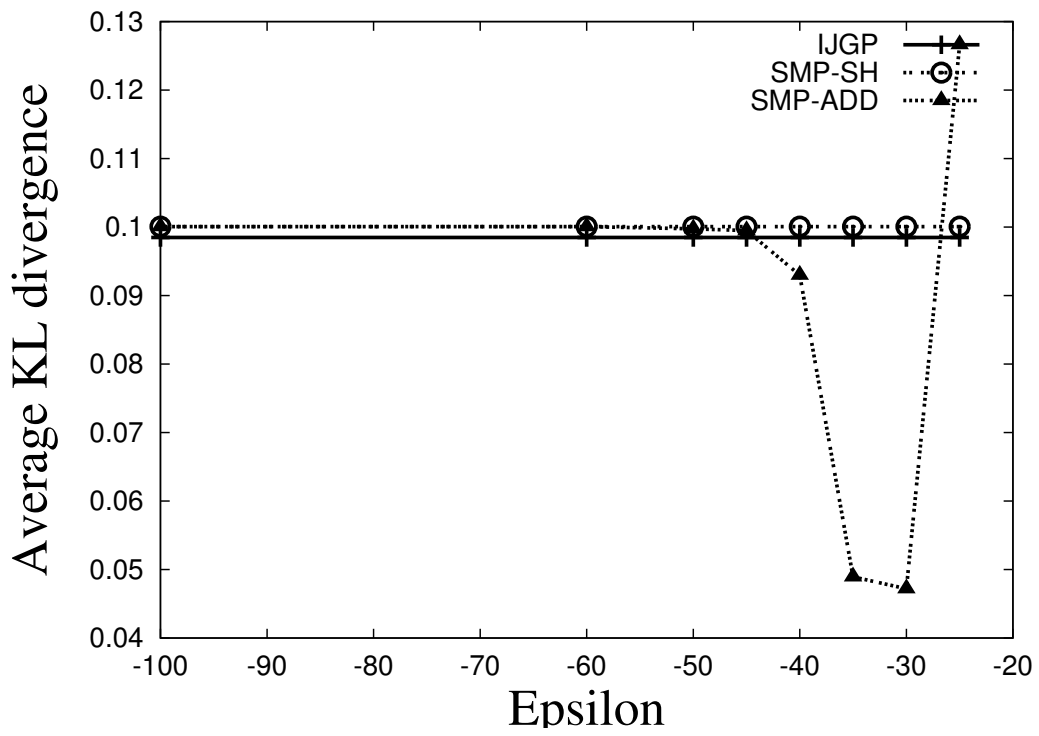

(a)

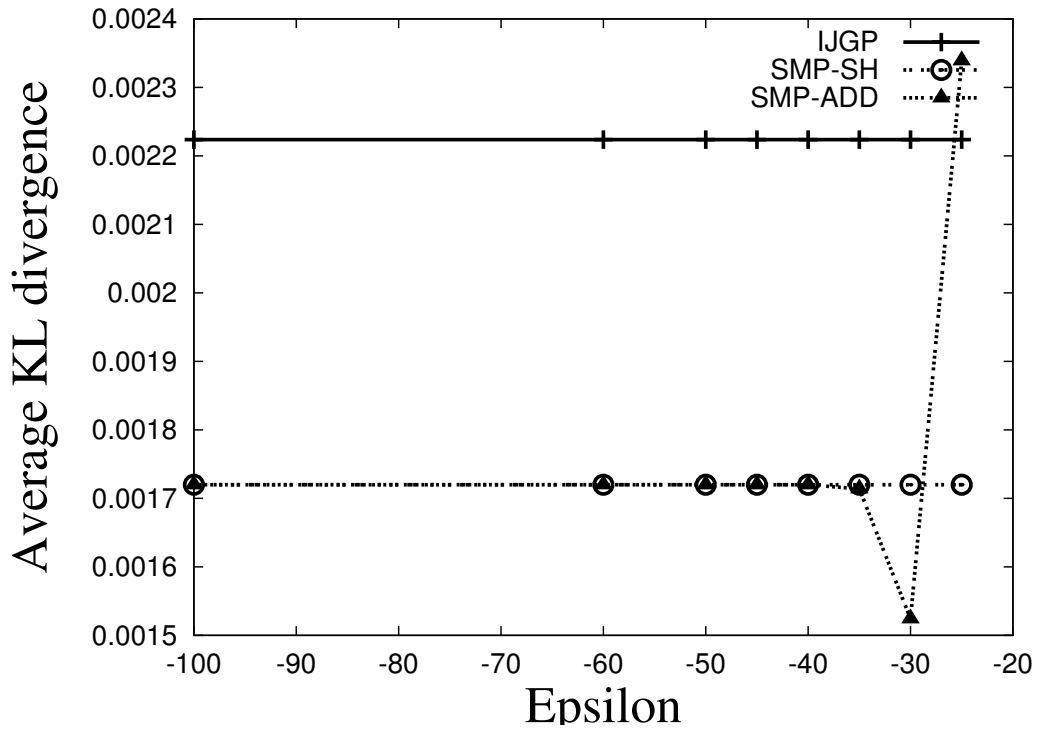

(b)

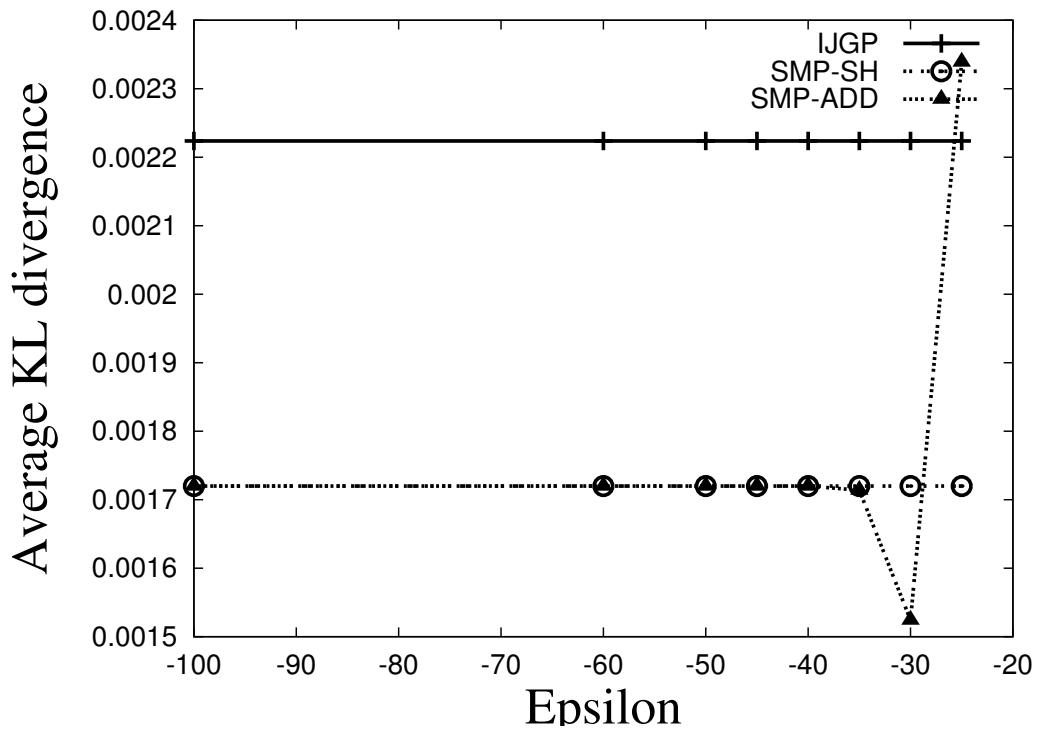

(c)

Figure 3: Impact of varying the quantization parameter $\epsilon$, keeping the i-bound and $k$ constant for (a) $20 \times 20$ Ising model with 10 evidence nodes, $i = 6$ and $k = 2^5$, (b) Block coding instance with 255 variables and 511 functions, $i = 12$ and $k = 2^{16}$ and (c) Logistics instance with 1413 nodes and 29487 functions, $i = 12$ and $k = 2^{16}$.

marginal estimation categories in the 2010 UAI competition [8]. We experimented with instances from three benchmark domains: (i) Ising models (these instances are available from the PASCAL 2011 probabilistic inference challenge), (ii) linear block coding (these instances available from the UAI 2008 evaluation), and (iii) logistic planning (these instances are available from the authors of Cachet [25]). Ising models have no determinism, the linear

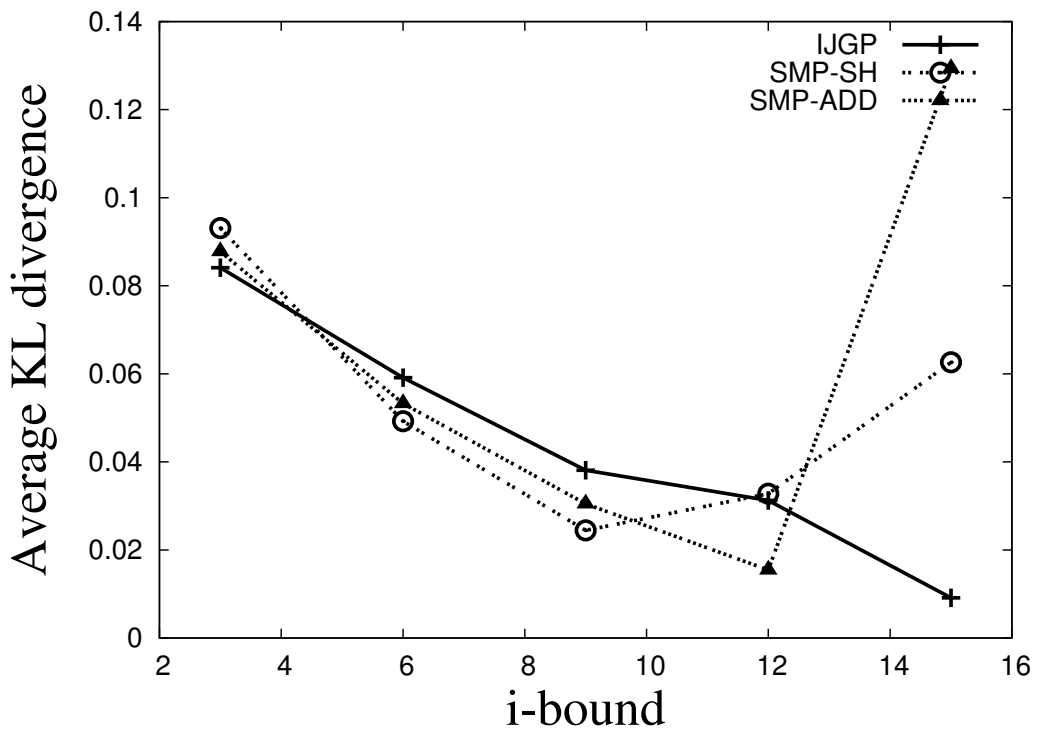

(a)

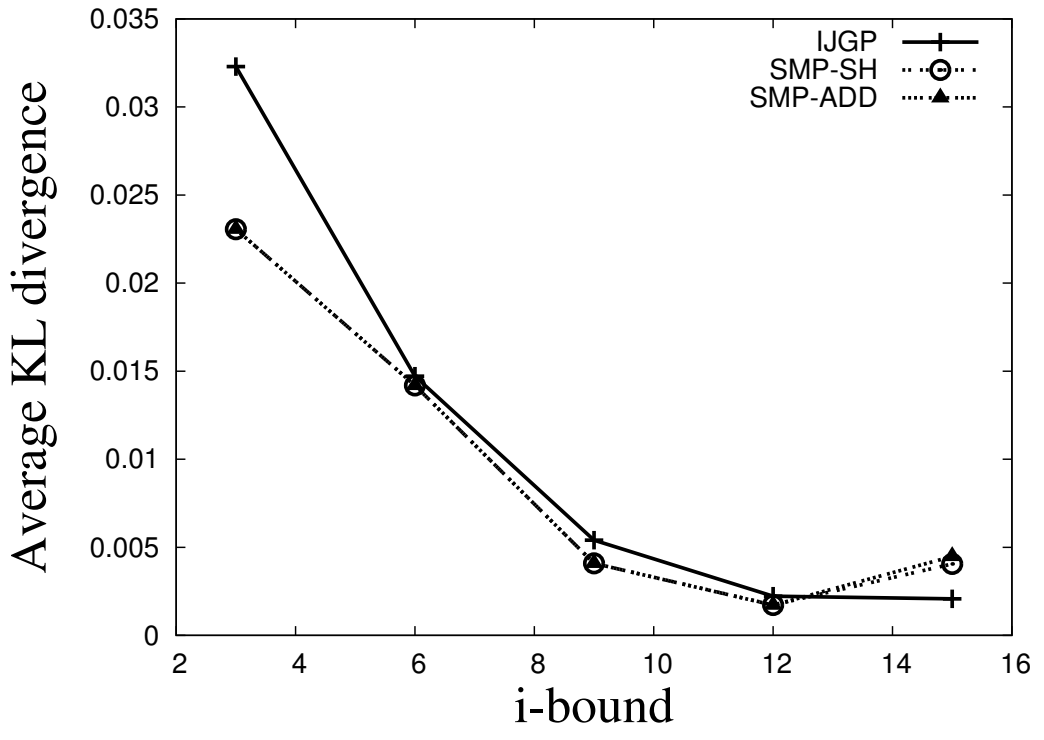

(b)

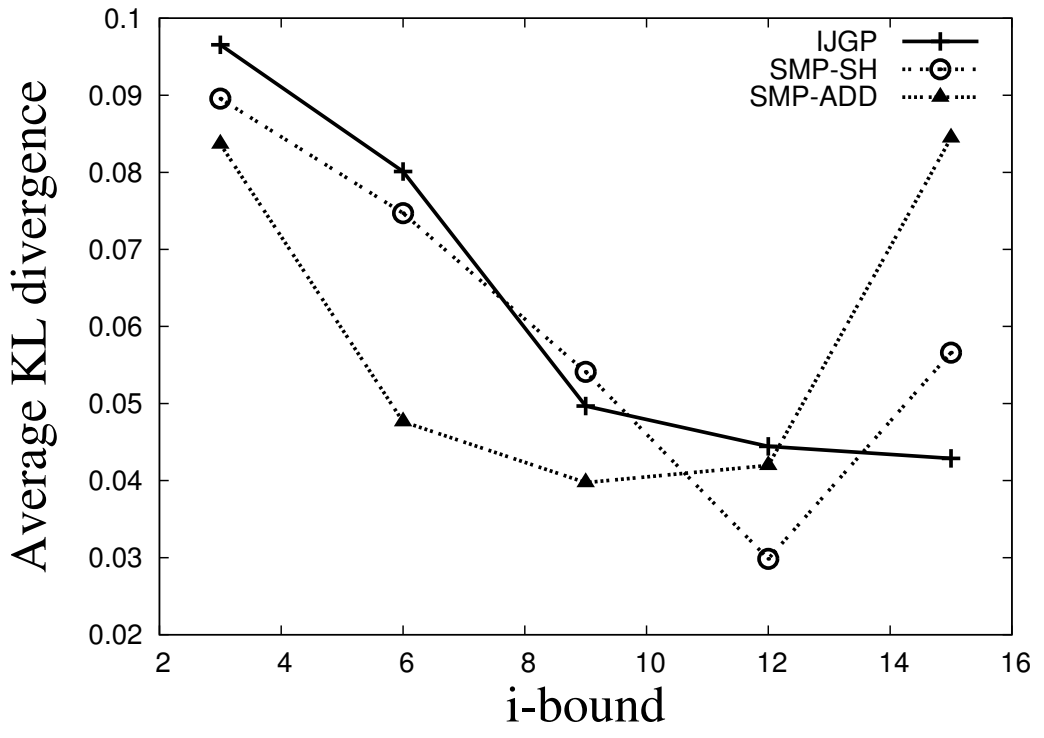

(c)

Figure 4: Impact of varying the i-bound, keeping $k$ and $\epsilon$ constant for (a) $20 \times 20$ Ising model with 5 evidence nodes, $k = 2^{14}$ and $\epsilon = 2^{-30}$, (b) Block coding instance with 255 variables and 511 functions, $k = 2^{18}$ and $\epsilon = 2^{-35}$ and (c) Logistics instance with 1413 nodes and 29487 functions, $k = 2^{16}$ and $\epsilon = 2^{-35}$.

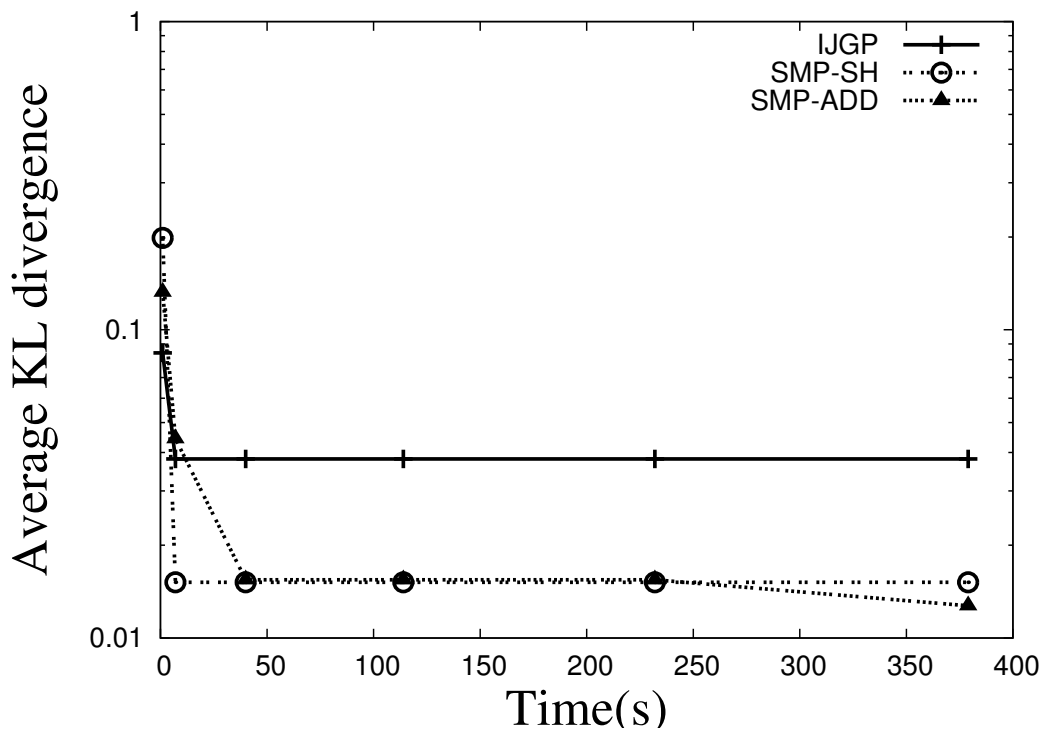

(a)

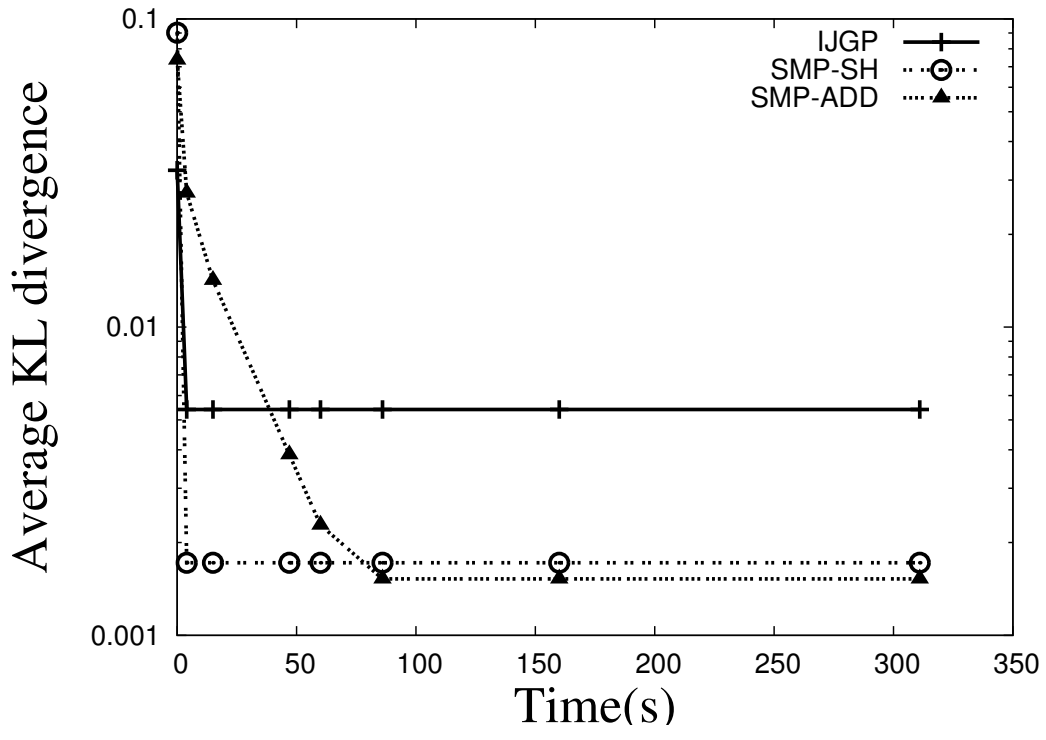

(b)

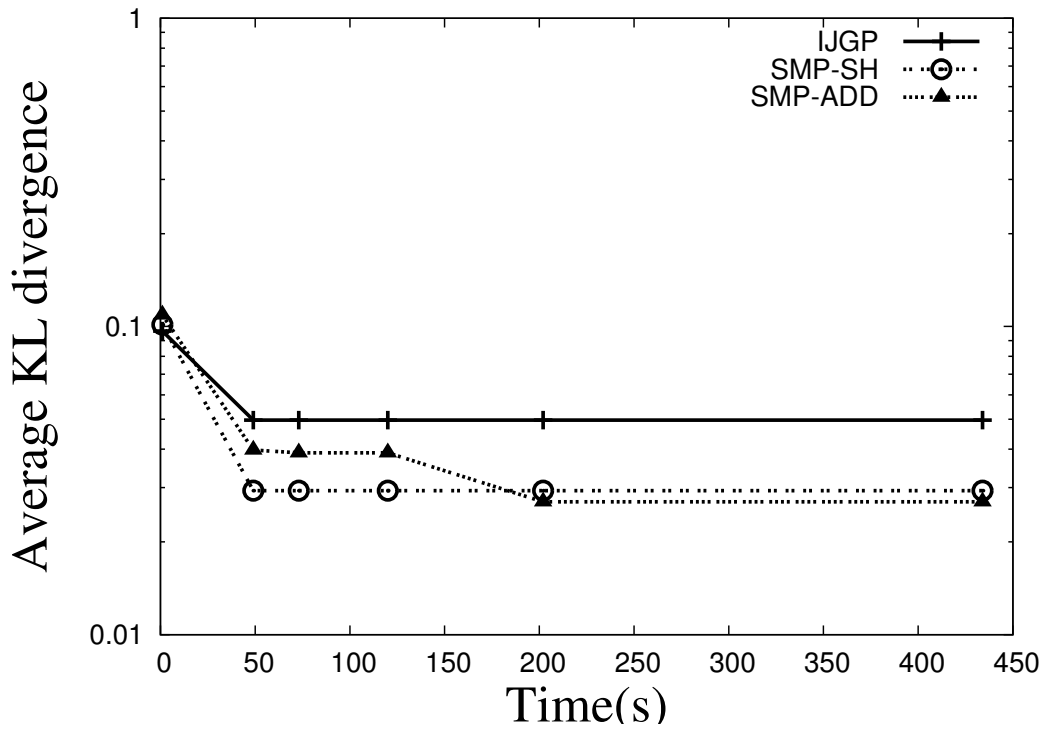

(c)

Figure 5: KL-divergence as a function of time (a) $21 \times 21$ Ising model with 5 evidence nodes, (b) Block coding instance with 255 variables and 511 functions and (c) Logistics instance with 1413 nodes and 29487 functions.

block coding networks have determinism and CSI while the logistic planning instances have determinism but no CSI.

We used the CUDD package [29] to implement ADDs. For a fair comparison, we constructed the cluster graphs for SMP using the same approach used by IJGP (see [19] for details). In IJGP, the complexity of inference is controlled by bounding the number of variables in each cluster by an integer parameter $i$, called its i-bound. We varied the i-bound from 3 to 15 in increments of 3. For the SMP algorithm, we varied $\epsilon$ (which controls quantization) from $2^{-20}$ to $2^{-100}$ and $k$ (number of samples) from $2^5$ to $2^{20}$. We used Gibbs sampling on the Ising models and importance sampling on the other two for generating samples (This is because Gibbs sampling does not converge in presence of determinism). We ran each algorithm until convergence or until 15 minutes or until it exceeded a memory bound of 512 MB, whichever was earlier. We chose these values because the benchmarks can be solved exactly in roughly 1 hour of cpu time using up to 8 GB of RAM. We used av-erage KL divergence between the exact and approximate marginal distribution to measure accuracy.

For lack of space, we only show a fraction of our results in Figures 2-5. SMP-SH and SMP-ADD denote the sparse hash table based and ADD based implementation of SMP respectively. Note that each point in each figure denotes an average over 10 random runs of IJGP, SMP-SH and SMP-ADD respectively.

Figure 2 shows the impact of varying the number of samples $k$, keeping the i-bound and $\epsilon$ constant for three instances, one from each domain. As expected, the accuracy of SMP-SH and SMP-ADD improves with more samples.

Figure 3 shows the impact of varying the quantization parameter $\epsilon$, keeping the i-bound and $k$ constant for three instances. In all three cases, we clearly see the bias versus variance trade-off; as we decrease $\epsilon$, the accuracy first improves and then reduces before stabilizing to a fixed point.

Figure 4 shows the impact of varying the i-bound, keeping $\epsilon$ and $k$ constant for three instances. Again in all three cases, we clearly see the bias versus variance trade-off; as we increase the cluster size (i-bound), the accuracy improves until a certain i-bound after which it starts decreasing.

Figure 5 shows the accuracy of various schemes as a function of time. For each time-point, we select the parameters that yield the best accuracy for each of the three methods. SMP-ADD is more accurate than SMP-SH which in turn is more accurate than IJGP. Note the log-scale on the Y-axis and therefore there is an order of magnitude difference.

# 7 RELATED WORK

Our work is related to the work on structured region graphs (SRGs) by Welling et al. [33]. In it, the authors showed that Yedidia et al.'s [34] generalized belief propagation algorithm *morphs* into the EP algorithm [20] if each message and cluster potential in the region graph is approximated by a product of tractable tabular functions. Our work is different from Welling et al.'s work in that we propose to use structured representations which are often more compact than the product of tabular functions representation. Also, unlike our formulation, there is no straight-forward way of introducing and exploiting determinism and CSI in the SRG formalism.

Our work is also related to the work on non-parametric BP by Sudderth et al. [31] and particle BP by Ihler and McAllester [14], in which the authors propose to represent BP messages using samples or particles. However, unlike our work, these approaches do not exploit structured representations and do not utilize both CSI and determinism. Also, they do not investigate bias versus variance trade-offs as we do. Our work connects particle BP with EP, yielding a more unified perspective.

Another related work is that of [7, 12, 23] who perform sampling based inference on junction trees. The main idea in these papers is to perform message passing on a junction tree by substituting messages which are too hard to compute exactly by their sampling-based approximations. Unlike our work, however, they do not perform message passing over arbitrary cluster graphs. This is problematic because as we showed, for a small sample size, junction trees will have low bias but high variance and as a result they will likely yield inaccurate results.

Finally, our work is related to the work on approximation by quantization (ABQ) by Gogate and Domingos [10]. Unlike ABQ which only introduces CSI, we propose to introduce both CSI and determinism which as we show often yields better accuracy in practice.

# 8 SUMMARY AND FUTURE WORK

In this paper, we proposed *structured message passing*, a unifying approach for taking advantage of structured representations. We investigated the use of two structured representations within this framework: algebraic decision diagrams (ADDs) and sparse hash tables. ADDs are useful in the presence of CSI and/or determinism while sparse tables are useful only in the presence of determinism. Therefore, in order to fully utilize the power of ADDs and sparse tables, we proposed a new algorithm that artificially introduces CSI via quantization and determinism via sampling. Our new algorithm is quite powerful and includes the cluster graph BP algorithm, the EP algorithm and the particle BP algorithm as special cases. Our algorithm introduces several bias versus variance tradeoffs. We investigated these tradeoffs both theoretically and empirically and showed that our new algorithm is superior to state-of-the-art approaches such as Iterative Join Graph Propagation [19].

Future work includes: applying our algorithm to continuous and hybrid discrete/continuous graphical models; using other structured representations such as mixture models and Affine ADDs [26] within SMP; combining SMP with lifted inference (cf. [11]); using our algorithm for weight learning; developing automatic tuning strategies for finding the right balance between bias and variance; etc.

**Acknowledgements**

This research was partly funded by ARO grant W911NF-08-1-0242, AFRL contracts FA8750-09-C-0181 and FA8750-13-2-0019, NSF grant IIS-0803481, and ONR grant N00014-12-1-0312. The views and conclusions contained in this document are those of the authors and should not be interpreted as necessarily representing the official policies, either expressed or implied, of ARO, AFRL, NSF, ONR, or the United States Government.